\pdfoutput=1 
\documentclass[a4paper,abstract]{scrartcl}
\usepackage{authblk}
\usepackage{times} 
\usepackage{helvet}
\usepackage{courier} 
\usepackage[hyphens]{url} 
\usepackage{graphicx} 
\usepackage{natbib}

\title{\Large Re-conceptualising the Language Game Paradigm in the Framework of Multi-Agent Reinforcement Learning}

\author[1,2]{Paul Van Eecke}
\author[1]{Katrien Beuls}
\affil[1]{Artificial Intelligence Laboratory, Vrije Universiteit Brussel \protect\\ Pleinlaan 2, 1050 Brussels, Belgium}
\affil[2]{ ITEC, imec research group at KU Leuven \protect\\ E. Sabbelaan 53, 8500 Kortrijk, Belgium }

\date{\small \textit{This paper was accepted for presentation at the 2020 AAAI Spring Symposium `Challenges and Opportunities for Multi-Agent Reinforcement Learning' after a double-blind reviewing process.}}

\begin{document}

\maketitle

\begin{abstract} 
\noindent
In this paper, we formulate the challenge of re-conceptualising the language game experimental paradigm in the framework of multi-agent reinforcement learning (MARL). If successful, future language game experiments will benefit from the rapid and promising methodological advances in the MARL community, while future MARL experiments on learning emergent communication will benefit from the insights and results gained from language game experiments. We strongly believe that this cross-pollination has the potential to lead to major breakthroughs in the modelling of how human-like languages can emerge and evolve in multi-agent systems.
\end{abstract}

\noindent 

\section{Introduction}
Learning emergent communication has become a topic of great interest in the broader AI community, as achieving robust, flexible and adaptive agent-agent and human-agent communication forms a key precondition for building truly intelligent systems \citep{mikolov2016roadmap}.  Multi-agent reinforcement learning (MARL) forms a natural framework for learning emergent communication, given its adequacy to model, to a large extent, the conditions under which human languages emerge and evolve. The MARL framework has as a consequence been adopted in a number of influential papers on emergent communication, tackling a wide variety of tasks, including visual question answering \citep{das2017learning}, solving puzzles \citep{foerster2016learning}, negotiation \citep{cao2018emergent}, reference \citep{lazaridou2016multi}, navigation \citep{sukhbaatar2016learning,bogin2018emergence,mordatch2018emergence}, and coordination in self-driving cars \citep{resnick2018vehicle}. The focus of these experiments is typically on learning emergent languages that are effective at solving the task at hand, which explains that the experimental conditions widely vary and sometimes seem far removed from how human languages have emerged and continue to evolve. While this is not a problem in itself, it has important repercussions on the linguistic systems that emerge, and on the operational deployability of the models. For example, agents are in these experiments often not autonomous\footnote{By autonomous agents, we mean agents that sense and act through their own sensors and actuators, make their own decisions, and are not under any form of central control.}, which poses problems when they are deployed in situations where they need to learn to communicate with agents that are not under the same central control (including human interlocutors), or do not share the same hardware or software architecture. 

Outside the MARL community, emergent communication is most prominently studied using the language game experimental paradigm \citep{steels1995self,steels2012experiments,beuls2013agent}. One of the key defining properties of this paradigm is that the circumstances under which emergent communication is modelled resemble as much as possible those under which human languages emerge. These circumstances, of which we would argue that many are in line with the MARL framework and none are fundamentally incompatible, include the following:

\begin{itemize}
\item Languages emerge and evolve in \textbf{a multi-agent setting}, namely in a population of agents that participate in situated communicative interactions.
\item Agents are \textbf{autonomous} and communicate through language. They possess no mind-reading or broadcasting capabilities.
\item Communicative interactions are \textbf{local} and learning is \textbf{de-centralised}. Only those agents that participate in an interaction can exploit its outcome for learning.
\item Communicative interactions are \textbf{goal-oriented}. They serve a communicative purpose and can as such succeed or fail.
\item The emerged languages are \textbf{shaped by past successes and failures in communication}.
\end{itemize}

We strongly believe that a cross-pollination between the language game paradigm and the MARL framework has the potential to lead to major breakthroughs in the modelling of how human-like languages can emerge and evolve in multi-agent systems. Future language game experiments could benefit from the rapid and promising methodological advances in the MARL community, while future MARL experiments on learning emergent communication could benefit from the insights and results gained from language game experiments.  

Therefore, in this paper, we formulate the challenge of re-conceptualising the language game experimental paradigm in the framework of multi-agent reinforcement learning. We first briefly introduce the language game methodology and provide a brief overview of prior experiments. Then, we provide a tentative overview of important challenges that might arise when bridging the gap between both paradigms. We hope that this will open a fruitful discussion between the MARL and language game communities, which can in turn lead to valuable collaborations.

\section{The Language Game Paradigm}

The language game paradigm\footnote{See \cite{steels2012self} for a brief introduction.} views language as an evolutionary system that emerges through the communicative interactions of language users and is shaped by the evolutionary processes of variation and selection. These processes take place within the linguistic system itself (rather than in the genes of the language users), on the level of concepts, words, grammar and discourse. Sources of variation mainly stem from the creativity and problem-solving capabilities of the language users, while the main selective pressures are success in communication and a reduction of processing effort. 

In terms of methodology, the language game paradigm employs multi-agent simulations for modelling the emergence and evolution of human-like languages. Such a simulation takes the form of a series of communicative interactions between autonomous agents in a population. A typical experiment proceeds as follows. At the beginning of each interaction, two agents are selected from the population and are assigned the role of either speaker or hearer. The agents are placed in a particular scene and need to successfully communicate to solve a given task, which often consists in referring to objects or events that they observe in the scene. The agents are equipped with mechanisms for inventing and adopting linguistic means (e.g. words, concepts or grammatical structures) that can be needed to achieve communicative success. After each interaction, the speaker provides feedback to the hearer about the outcome of the task. This allows the hearer to learn in the case that the agents did not reach communicative success. Additionally, both agents align by rewarding and updating the linguistic means that were used in the case of a successful interaction, and punishing these in the case of a failed interaction. As more and more games are played, the agents in the population gradually converge on a shared language. The language of each individual agent has been shaped by the communicative interactions it participated in and is, therefore, well adapted to the task and the environment.

\begin{figure}
\centering
\includegraphics[width=\columnwidth]{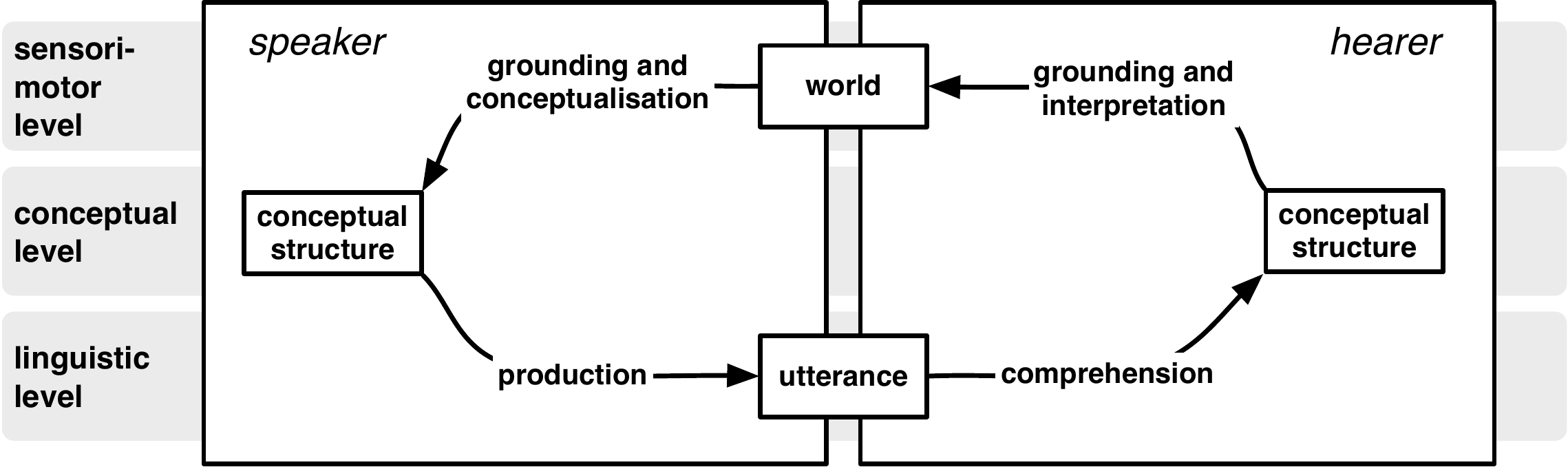}
\caption{The semiotic cycle depicts the sensori-motor, conceptual and linguistic processes that a language game involves for a speaker agent (left) and a hearer agent (right).}
\label{fig:semiotic}
\end{figure}

During a communicative interaction, the speaker and hearer agents go through the different processes depicted in Figure \ref{fig:semiotic}. Both agents are situated in the same physical or simulated world, which they perceive through their sensori-motor system. The speaker agent maps its sensori-motor experiences to meaningful concepts and conceptual structures, abstracting away from the raw sensor values (\textit{grounding and conceptualisation}). These conceptual structures are then expressed as linguistic utterances (\textit{production}). The hearer agent perceives the utterances and uses its own linguistic system to construct conceptual structures (\textit{comprehension}), which it then interprets with respect to the world (\textit{grounding and interpretation}). 

The ultimate goal of language game experiments is to find adequate invention, adoption and alignment mechanisms that allow a population of agents to self-organise a conceptual and linguistic system that allows them to communicate in order to solve an open-ended set of tasks in an ever-changing environment.

\section{A Brief Overview of Prior Experiments}

Experiments in emergent communication using the language game methodology can be grouped into three main categories, based on the type of linguistic system that is modelled. The earliest and most widely known experiments investigated the emergence of vocabularies. Later experiments focused on the emergence and co-evolution of concepts and words, moving away from purely symbolic to continuous observations. Currently, most research within the language game paradigm centers around modelling the emergence and co-evolution of complex conceptual and grammatical structures. 

The first experiments on emergent communication, for which the language game paradigm was originally conceived, investigated the \textbf{emergence of vocabularies} in the form of inventories that consist of conventionalised mappings between word forms and objects or symbolic object properties \citep{steels1995self}. These experiments have become known as naming game experiments.  Naming games have been extensively studied in the literature from many different angles, including scaling laws \citep{baronchelli2006sharp}, convergence proofs \citep{devylder2006reach}, alignment strategies \citep{wellens2012adaptive},  social network topologies \citep{liu2009naming,lipowska2012naming}, and learning strategies \citep{schueller2016active}. The large body of literature around the naming game has convincingly shown how a population of agents can self-organise through purely local interactions a shared vocabulary that allows them to communicate using individual words.
 
The second wave of experiments originated from these naming game experiments, and extended their scope to the \textbf{emergence of grounded concepts}. These experiments study the mechanisms that make it possible for a population of agents to self-organise a conceptual system and vocabulary through local communicative interactions. Such a system forms an abstraction layer that maps between raw, continuous sensori-motor observations and symbols that correspond to meaningful distinctions in the task and environment. Experiments on emergent conceptual systems and vocabularies have been conducted in different domains, including color \citep{steels2005coordinating,puglisi2008cultural}, vowel systems \citep{de2000self,oudeyer2001coupled}, spatial relations \citep{spranger2012coevolution} and action language \citep{steels2012emergent}, as well as in abstract worlds \citep{spranger2016referential}. While these experiments have yielded a basic understanding of the mechanisms involved in the emergence of concepts and words, many open questions remain, especially when it comes to learning domain-general concepts and concepts that are not limited to visually observable features.  

The third wave of language game experiments goes beyond individual words and concepts, by studying the \textbf{emergence of conceptual structures and grammatical expressions}. These experiments investigate how more complex semantic structures, along with linguistic systems that express them, can emerge and evolve in a population of autonomous agents. One approach to this problem consists in modelling how complex compositional and hierarchical semantic structures can emerge along with morpho-syntactic patterns that reflect their composition. Experiments that adopt this approach have been carried out in the domains of, amongst others, spatial relations (e.g. `the block left of the ball') \citep{spranger2016evolution}, quantifiers (e.g. `many large boxes') \citep{pauw2016embodied} and logical operators (e.g. `either large or red, but not both') \citep{sierra2018agent}. The second approach studies how morpho-syntactic structures might arise to dampen the referential ambiguity that arises when longer utterances are used. This is typically achieved by means of emergent word order \citep{steels2015ambiguity,vaneecke2018generalisation} or agreement marking systems \citep{beuls2011simulating,beuls2013agent,vantrijp2016evolution,lestrade2016emergence}. Although this range of experiments shows great promise for the application of the language game methodology to emergent conceptual and grammatical systems, this domain still remains to a large extent uncharted territory. 

\section{Challenges for MARL}

Many of the central ideas that underlie the language game paradigm are also characteristic of the MARL framework as applied to experiments in emergent communication. First and foremost, both methodologies make use of multi-agent simulations to investigate how a population of agents can learn to communicate through task-based communicative interactions. Second, the main forces driving the dynamics of the simulation are the rewarding of the agent's language use in  the case of a successful interaction and the punishing of its language use in the case of a failed interaction. Finally, the languages that emerge can be human languages that were learnt in a tutor-learner scenario or artificial languages that do not exist outside the simulation. 

There are however other aspects of the language game paradigm that are more challenging to fully operationalise within the MARL framework, but that are often considered to be desirable properties of emergent communication experiments and have the potential to lead to the emergence of more human-like languages. These aspects include the following:

\begin{itemize}
\item Agents should be \textbf{fully autonomous}, in the sense that they make their own decisions and are not subject to any form of central control. They should not have mind-reading or broadcasting capabilities and interact only with the world and each other through their own sensors and actuators. This is necessary to ensure that the languages can emerge in populations of heterogeneous agents, which might not share the same hardware and software architectures.

\item The communicative interactions should be \textbf{local} and only accessible to the agents that participate. Consequently, this means that learning should be \textbf{decentralised}, so that the languages emerge through self-organisation (i.e. a global system arising from purely local interactions). Such decentralised, self-organising systems are known to be able to self-repair substantial perturbations, a form of robustness that is necessary for modelling the emergence and evolution of truly human-like languages.

\item The agents in the population should be able to take up the roles of \textbf{both speaker and hearer} and their comprehension and production processes should be integrated. The agents should be able to express the concepts, words and grammatical structures that they have learned in the hearer role, and be able to understand the utterances that they produced in the speaker role.

\item The emergent languages should be as \textbf{transparent} as possible, and at least to the extent that their communicative function can be properly evaluated \citep{lowe2019pitfalls}.

\item The emergent languages should be \textbf{flexible and adaptive} to changes in the tasks and environment of the agents. It should be avoided at all cost that a different language needs to emerge when changes in the tasks and environment occur.

\item The linguistic inventories that contain representations of concepts, words and grammatical structures should be \textbf{dynamically expandable}, so that new words, concepts and grammatical structures can be introduced when needed.

\end{itemize}

\section{Conclusion and Outlook}

Multi-agent reinforcement learning forms a natural framework for conducting experiments on learning emergent communication, and has been adopted as a methodology of choice in many of the most influential recent papers on the topic. However, when it comes to modelling the emergence of robust, flexible and adaptive human-like languages, a number of important limitations remain. Most of these limitations relate to the experimental set-ups that are used and more in particular to the circumstances under which the languages emerge and evolve. We have therefore formulated the challenge of re-conceptualising the language game experimental paradigm -- of which modelling circumstances that resemble as closely as possible those under which human languages emerge has always been one of the main concerns --  in the MARL framework, and have identified a number of key aspects that will need to be operationalised for achieving this goal. If successful, future language game experiments will be able to benefit from the rapid and promising methodological advances in the MARL community, while future MARL experiments on learning emergent communication will  be able  to benefit from the insights gained from language game experiments, thereby yielding more human-like emergent languages.

We hope that this paper will open a fruitful discussion between the MARL and language game communities, which can in turn lead to valuable collaborations that will push forward the state of the art in learning emergent communication.

\setlength{\bibsep}{1pt plus 0.3ex}
\bibliographystyle{apalike}
\bibliography{../../../../ehaidocs/Bibliography/ehai_bibliography}

\begin{thebibliography}{}

\bibitem[Baronchelli et~al., 2006]{baronchelli2006sharp}
Baronchelli, A., Felici, M., Loreto, V., Caglioti, E., and Steels, L. (2006).
\newblock Sharp transition towards shared vocabularies in multi-agent systems.
\newblock {\em Journal of Statistical Mechanics: Theory and Experiment},
  2006(06):P06014.

\bibitem[Beuls and H{\"o}fer, 2011]{beuls2011simulating}
Beuls, K. and H{\"o}fer, S. (2011).
\newblock Simulating the emergence of grammatical agreement in multi-agent
  language games.
\newblock In {\em Proceedings of the Twenty-Second International Joint
  Conference on Artificial Intelligence}, pages 61--66.

\bibitem[Beuls and Steels, 2013]{beuls2013agent}
Beuls, K. and Steels, L. (2013).
\newblock Agent-based models of strategies for the emergence and evolution of
  grammatical agreement.
\newblock {\em PloS one}, 8(3):e58960.

\bibitem[Bogin et~al., 2018]{bogin2018emergence}
Bogin, B., Geva, M., and Berant, J. (2018).
\newblock Emergence of communication in an interactive world with consistent
  speakers.
\newblock {\em arXiv preprint arXiv:1809.00549}.

\bibitem[Cao et~al., 2018]{cao2018emergent}
Cao, K., Lazaridou, A., Lanctot, M., Leibo, J.~Z., Tuyls, K., and Clark, S.
  (2018).
\newblock Emergent communication through negotiation.
\newblock {\em arXiv preprint arXiv:1804.03980}.

\bibitem[Das et~al., 2017]{das2017learning}
Das, A., Kottur, S., Moura, J.~M., Lee, S., and Batra, D. (2017).
\newblock Learning cooperative visual dialog agents with deep reinforcement
  learning.
\newblock In {\em Proceedings of the IEEE International Conference on Computer
  Vision}, pages 2951--2960.

\bibitem[{de Boer}, 2000]{de2000self}
{de Boer}, B. (2000).
\newblock Self-organization in vowel systems.
\newblock {\em Journal of phonetics}, 28(4):441--465.

\bibitem[{De Vylder} and Tuyls, 2006]{devylder2006reach}
{De Vylder}, B. and Tuyls, K. (2006).
\newblock How to reach linguistic consensus: A proof of convergence for the
  naming game.
\newblock {\em Journal of theoretical biology}, 242(4):818--831.

\bibitem[Foerster et~al., 2016]{foerster2016learning}
Foerster, J., Assael, I.~A., de~Freitas, N., and Whiteson, S. (2016).
\newblock Learning to communicate with deep multi-agent reinforcement learning.
\newblock In {\em Advances in Neural Information Processing Systems}, pages
  2137--2145.

\bibitem[Lazaridou et~al., 2016]{lazaridou2016multi}
Lazaridou, A., Peysakhovich, A., and Baroni, M. (2016).
\newblock Multi-agent cooperation and the emergence of (natural) language.
\newblock {\em arXiv preprint arXiv:1612.07182}.

\bibitem[Lestrade, 2016]{lestrade2016emergence}
Lestrade, S. (2016).
\newblock The emergence of argument marking.
\newblock In Roberts, S., Cuskley, C., McCrohon, L., Barcel\'{o}-Coblijn, L.,
  Feh\'{e}r, O., and Verhoef, T., editors, {\em The Evolution of Language:
  Proceedings of the 11th International Conference (EVOLANGX11)}. Online at
  \url{http://evolang.org/neworleans/papers/36.html}.

\bibitem[Lipowska and Lipowski, 2012]{lipowska2012naming}
Lipowska, D. and Lipowski, A. (2012).
\newblock Naming game on adaptive weighted networks.
\newblock {\em Artificial Life}, 18(3):311--323.

\bibitem[Liu et~al., 2009]{liu2009naming}
Liu, R.-R., Jia, C.-X., Yang, H.-X., and Wang, B.-H. (2009).
\newblock Naming game on small-world networks with geographical effects.
\newblock {\em Physica A: Statistical Mechanics and its Applications},
  388(17):3615--3620.

\bibitem[Lowe et~al., 2019]{lowe2019pitfalls}
Lowe, R., Foerster, J., Boureau, Y.-L., Pineau, J., and Dauphin, Y. (2019).
\newblock On the pitfalls of measuring emergent communication.
\newblock In {\em Proceedings of the 18th International Conference on
  Autonomous Agents and MultiAgent Systems}, pages 693--701.

\bibitem[Mikolov et~al., 2016]{mikolov2016roadmap}
Mikolov, T., Joulin, A., and Baroni, M. (2016).
\newblock A roadmap towards machine intelligence.
\newblock In {\em Proceedings of the International Conference on Intelligent
  Text Processing and Computational Linguistics}, pages 29--61.

\bibitem[Mordatch and Abbeel, 2018]{mordatch2018emergence}
Mordatch, I. and Abbeel, P. (2018).
\newblock Emergence of grounded compositional language in multi-agent
  populations.
\newblock In {\em Proceedings of the Thirty-Second AAAI Conference on
  Artificial Intelligence}, pages 1495--1502.

\bibitem[Oudeyer, 2001]{oudeyer2001coupled}
Oudeyer, P.-Y. (2001).
\newblock Coupled neural maps for the origins of vowel systems.
\newblock In {\em International Conference on artificial Neural Networks},
  pages 1171--1176. Springer.

\bibitem[Pauw and Hilferty, 2016]{pauw2016embodied}
Pauw, S. and Hilferty, J. (2016).
\newblock Embodied cognitive semantics for quantification.
\newblock {\em Belgian Journal of Linguistics}, 30(1):251--264.

\bibitem[Puglisi et~al., 2008]{puglisi2008cultural}
Puglisi, A., Baronchelli, A., and Loreto, V. (2008).
\newblock Cultural route to the emergence of linguistic categories.
\newblock {\em Proceedings of the National Academy of Sciences},
  105(23):7936--7940.

\bibitem[Resnick et~al., 2018]{resnick2018vehicle}
Resnick, C., Kulikov, I., Cho, K., and Weston, J. (2018).
\newblock Vehicle communication strategies for simulated highway driving.
\newblock {\em arXiv preprint arXiv:1804.07178}.

\bibitem[Schueller and Oudeyer, 2016]{schueller2016active}
Schueller, W. and Oudeyer, P.-Y. (2016).
\newblock Active control of complexity growth in naming games: Hearer's choice.
\newblock In Roberts, S., Cuskley, C., McCrohon, L., Barcel\'{o}-Coblijn, L.,
  Feh\'{e}r, O., and Verhoef, T., editors, {\em The Evolution of Language:
  Proceedings of the 11th International Conference (EVOLANGX11)}. Online at
  \url{http://evolang.org/neworleans/papers/105.html}.

\bibitem[Sierra-Santib{\'a}{\~n}ez, 2018]{sierra2018agent}
Sierra-Santib{\'a}{\~n}ez, J. (2018).
\newblock An agent-based model of the emergence and evolution of a language
  system for boolean coordination.
\newblock {\em Autonomous Agents and Multi-Agent Systems}, 32(4):417--458.

\bibitem[Spranger, 2012]{spranger2012coevolution}
Spranger, M. (2012).
\newblock The co-evolution of basic spatial terms and categories.
\newblock In Steels, L., editor, {\em {E}xperiments in {C}ultural {L}anguage
  {E}volution}, number~3 in Advances in Interaction Studies, pages 111--141.
  John Benjamins.

\bibitem[Spranger, 2016]{spranger2016evolution}
Spranger, M. (2016).
\newblock {\em The evolution of grounded spatial language}.
\newblock Language Science Press, Berlin.

\bibitem[Spranger and Beuls, 2016]{spranger2016referential}
Spranger, M. and Beuls, K. (2016).
\newblock Referential uncertainty and word learning in high-dimensional,
  continuous meaning spaces.
\newblock In {\em 2016 Joint IEEE International Conference on Development and
  Learning and Epigenetic Robotics (ICDL-EpiRob)}, pages 95--100. IEEE.

\bibitem[Steels, 1995]{steels1995self}
Steels, L. (1995).
\newblock A self-organizing spatial vocabulary.
\newblock {\em Artificial life}, 2(3):319--332.

\bibitem[Steels, 2012a]{steels2012experiments}
Steels, L., editor (2012a).
\newblock {\em Experiments in cultural language evolution}.
\newblock John Benjamins, Amsterdam.

\bibitem[Steels, 2012b]{steels2012self}
Steels, L. (2012b).
\newblock Self-organization and selection in cultural language evolution.
\newblock In Steels, L., editor, {\em Experiments in Cultural Language
  Evolution}, pages 1--37. John Benjamins, Amsterdam.

\bibitem[Steels and Belpaeme, 2005]{steels2005coordinating}
Steels, L. and Belpaeme, T. (2005).
\newblock Coordinating perceptually grounded categories through language: A
  case study for colour.
\newblock {\em Behavioral and brain sciences}, 28(4):469--488.

\bibitem[Steels and Casademont, 2015]{steels2015ambiguity}
Steels, L. and Casademont, E.~G. (2015).
\newblock Ambiguity and the origins of syntax.
\newblock {\em The Linguistic Review}, 32(1):37--60.

\bibitem[Steels et~al., 2012]{steels2012emergent}
Steels, L., Spranger, M., {van Trijp}, R., H{\"o}fer, S., and Hild, M. (2012).
\newblock Emergent action language on real robots.
\newblock In Steels, L. and Hild, M., editors, {\em Language grounding in
  robots}, pages 255--276. Springer, New York.

\bibitem[Sukhbaatar et~al., 2016]{sukhbaatar2016learning}
Sukhbaatar, S., Szlam, A., and Fergus, R. (2016).
\newblock Learning multiagent communication with backpropagation.
\newblock In {\em Advances in Neural Information Processing Systems}, pages
  2244--2252.

\bibitem[{Van Eecke}, 2018]{vaneecke2018generalisation}
{Van Eecke}, P. (2018).
\newblock {\em Generalisation and Specialisation Operators for Computational
  Construction Grammar and their Application in Evolutionary Linguistics
  Research}.
\newblock PhD thesis, Vrije Universiteit Brussel, Brussels: VUB Press.

\bibitem[{van Trijp}, 2016]{vantrijp2016evolution}
{van Trijp}, R. (2016).
\newblock {\em The evolution of case grammar}.
\newblock Language Science Press, Berlin.

\bibitem[Wellens, 2012]{wellens2012adaptive}
Wellens, P. (2012).
\newblock {\em Adaptive Strategies in the Emergence of Lexical Systems}.
\newblock PhD thesis, Vrije Universiteit Brussel, Brussels: VUB Press.

\end{thebibliography}

\end{document}